\documentclass[a4paper,fleqn]{cas-dc}

\usepackage[numbers]{natbib}

\def\tsc#1{\csdef{#1}{\textsc{\lowercase{#1}}\xspace}}
\tsc{WGM}
\tsc{QE}
\tsc{EP}
\tsc{PMS}
\tsc{BEC}
\tsc{DE}
\let\printorcid\relax
\begin{document}
\let\WriteBookmarks\relax
\def\floatpagepagefraction{1}
\def\textpagefraction{.001}
\shorttitle{Dual-Codebook Graph Collaborative Network}
\shortauthors{Hongtao Li et~al.}

\title [mode = title]{Atrial Fibrillation Detection with Arbitrary Leads via a Codebook-Based Reconstruction-Classification Framework} 

\author[1]{Hongtao Li}
\affiliation[1]{organization={Information Center, The People's Hospital of Baoan Shenzhen},
                % addressline={Baoan}, 
                city={Shenzhen},
%               citysep={}, % Uncomment if no comma needed between city and postcode
                postcode={518101}, 
                state={Guangdong},
                country={China}}

\author[2]{Jia Wei}
\affiliation[2]{organization={School of Computer Science and Engineering, South China University of Technology},
                % addressline={XX}, 
                city={Guangzhou},
%               citysep={}, % Uncomment if no comma needed between city and postcode
                postcode={510006}, 
                state={Guangdong},
                country={China}}
\author[3]{Guoyao Li}
\affiliation[3]{organization={Department of Cardiology, The People's Hospital of Baoan Shenzhen},
                % addressline={Baoan}, 
                city={Shenzhen},
%               citysep={}, % Uncomment if no comma needed between city and postcode
                postcode={518101}, 
                state={Guangdong},
                country={China}}
\author[4]{Yuchen Lei}
\affiliation[4]{organization={Institute for Advanced Study, Shenzhen University},
                % addressline={XX}, 
                city={Shenzhen},
%               citysep={}, % Uncomment if no comma needed between city and postcode
                postcode={518060}, 
                state={Guangdong},
                country={China}}
\author[1]{Guangnian Ma}
\author[1]{Jia Xiao}
\author[1]{Yuanjun Lai}
\author[1]{Shuzhen Lv}
\author[1]{Xueqiang Ouyang}
\corref{cor1}
\cortext[cor1]{Corresponding author}
\ead{xqoy.bahosp@outlook.com}

\begin{abstract}
\textbf{Background and Objective}: Reliable atrial fibrillation (AF) detection from electrocardiogram (ECG) signals remains challenging in real-world clinical settings due to variable lead configurations, cross-dataset domain shifts, and pervasive physiological and technical artifacts. So we develop a robust and generalizable deep learning model for accurate AF detection.\\
\textbf{Methods}: We propose the Dual-Codebook Graph Collaborative Network (DCGCNet), a novel end-to-end vector-quantized variational autoencoder that jointly performs AF classification and ECG reconstruction. DCGCNet introduces two key components: (1) a Local-Global Contrastive Module for learning noise-invariant representations, and (2) an Adaptive Codebook Vector Quantizer that dynamically refines codebook prototypes to better align with input data distributions, thereby preventing codebook collapse and enhancing generalization.\\
\textbf{Results}: DCGCNet achieves state-of-the-art performance in standard intra-dataset 12-lead evaluation and demonstrates exceptional cross-dataset generalization across seven diverse settings, consistently attaining AUC > 0.98 in all cases. Furthermore, it maintains high diagnostic accuracy under realistic noisy conditions, including baseline wander, powerline interference, and EMG artifacts.\\
\textbf{Conclusions}: DCGCNet establishes a new benchmark for robust, generalizable, and noise-resilient AF detection, showing strong potential for deployment in real-world clinical environments.
\end{abstract}

% \begin{highlights}
% \item First codebook-based learning framework for AF detection and ECG reconstruction.
% \item Dual codebooks disentangle rhythm and morphology for synergistic learning.
% \item Robust and generalizable AF detection under noise and cross-dataset shift settings.
% \item Multi-level interpretability via codebook usage and Grad-CAM.
% \end{highlights}
% \includegraphics[width=1.0\textwidth]{figs/abstract}

\maketitle

\section{Introduction}

Atrial Fibrillation (AF), the most prevalent cardiac arrhythmia in clinical practice, has increasingly become a critical global public health burden \cite{healthbruden}. Recent epidemiological investigations have demonstrated that the number of individuals affected by AF worldwide has exceeded sixty million, and its prevalence continues to rise steadily, primarily driven by the global aging population and the changing lifestyles of modern societies \cite{risesteadily}. AF is closely associated with a significantly elevated risk of severe complications, including ischemic stroke and heart failure, which not only compromise patients’ quality of life but also impose an enormous economic burden on healthcare systems across the globe \cite{complications}. Therefore, the development of efficient, accurate, and reliable automated AF detection technologies is of paramount importance for facilitating early clinical intervention, optimizing disease management, and ultimately improving long-term patient outcomes \cite{detection}.

In clinical workflows, the electrocardiogram (ECG) remains the gold standard for AF diagnosis. However, conventional AF detection relies heavily on manual visual interpretation by specialized cardiologists, which is time-consuming, labor-intensive, and prone to subjective bias, limiting its utility for large-scale screening and real-time monitoring. In recent years, Artificial Intelligence (AI), particularly deep learning algorithms, has achieved transformative breakthroughs in ECG analysis, offering a promising approach for automated AF detection \cite{automated}.

\begin{figure}
	\centering
	\includegraphics[width=1.0\columnwidth]{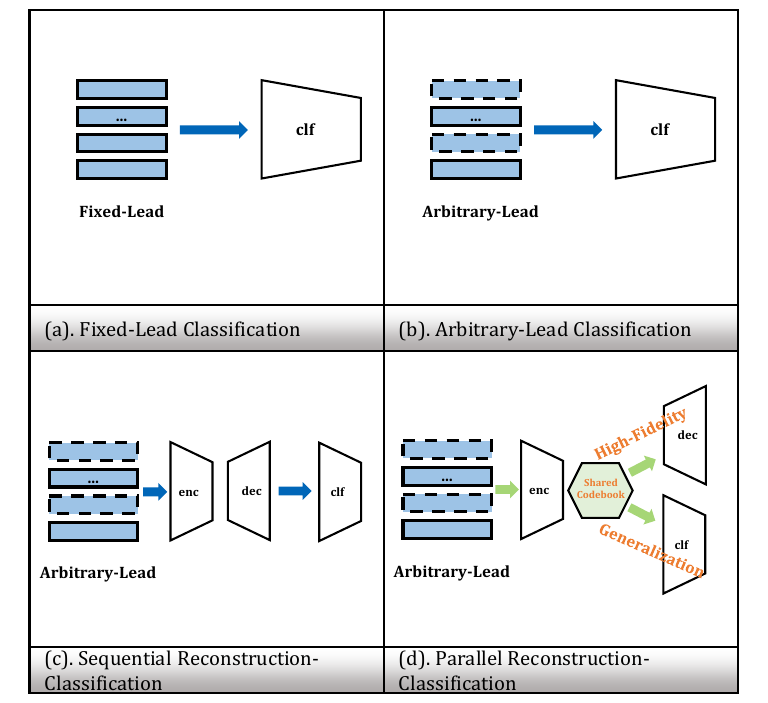}
	\caption{
        Evolution of AF detection frameworks and our motivation. 
        (a) Fixed-Lead models are inflexible to varying inputs. 
        (b) Arbitrary-Lead models suffer from information scarcity. 
        (c) Sequential frameworks decouple tasks, causing error propagation. 
        (d) Our parallel framework jointly optimizes both tasks for superior performance.
        }
	\label{fig:motivation}
\end{figure}

Existing AF detection methodologies of deep learning exhibit a clear evolution in addressing the challenge of variable ECG lead configurations, as illustrated in Figure~\ref{fig:motivation}. The initial approach, \textbf{Fixed-Lead Classification} (Figure~\ref{fig:motivation}(a)), operates under the assumption of a standardized input, typically a fixed set of high-quality leads. While this paradigm can yield excellent performance in controlled clinical settings, its fundamental limitation is a complete lack of flexibility; it fails catastrophically when deployed in real-world scenarios involving diverse or non-standard lead setups \cite{fixed1}\cite{fixed2}\cite{fixed3}.

To overcome this rigidity, research has shifted toward \textbf{Arbitrary-Lead Classification} (Figure~\ref{fig:motivation}(b)), which aims to develop models agnostic to the specific input leads. However, this direct approach is fundamentally limited by the incomplete diagnostic information inherent in non-standard or reduced-lead ECGs, as it lacks the comprehensive spatial perspective provided by a full 12-lead set, often resulting in suboptimal and unstable diagnostic performance \cite{arbitrary}.

A further approach, \textbf{Sequential Reconstruction Classification} (Figure~\ref{fig:motivation}(c)), bridges this gap by first reconstructing full standard 12-lead signals from arbitrary inputs via a dedicated network, before feeding the reconstruction into a separate classifier. Despite enhanced adaptability, this two-stage pipeline has a critical flaw: decoupled reconstruction and classification objectives. The reconstructor optimizes for signal fidelity, which may misalign with the classifier’s need for discriminative, diagnosis-relevant features, thereby risking the amplification of noise or the discarding of crucial diagnostic information \cite{recon1}\cite{recon2}\cite{recon3}.

To address these limitations, we propose a \textbf{Codebook-based Parallel Reconstruction-Classification Framework} (Figure~\ref{fig:motivation}(d)). The framework employs a shared codebook that jointly governs both objectives. This discrete representation is naturally suited to signal reconstruction, learning a compact set of prototypical ECG patterns for high-fidelity recovery from incomplete inputs \cite{vqvae}\cite{vqvaeapp}. Simultaneously, by projecting latent features onto this finite set of learned prototypes, the codebook implicitly regularizes the classifier, filtering out instance-specific noise and enhancing generalization \cite{dg}\cite{dgapp}.The two tasks are synergistically coupled through the shared codebook: reconstruction populates it with physiologically plausible structures, while classification shapes it to prioritize diagnostically discriminative features. This unified formulation ensures the reconstructed signal is both anatomically faithful and optimally informative for robust classification. The main contributions of this work are summarized as follows:

\textbf{1. We propose the first codebook-based joint learning framework for AF detection}, DCGCNet—an end-to-end VQ-VAE that unifies classification and reconstruction via two specialized codebooks: a \textit{rhythm codebook} for local discriminative patterns and a \textit{morphology codebook} for global signal fidelity, enabling synergistic representation learning.

\textbf{2. We introduce two tailored modules to boost robustness and generalization}: a \textit{Local-Global Contrastive Module} (LGCM) that enhances noise resilience through hierarchical contrastive learning on rhythm features, and an \textit{Adaptive Codebook Vector Quantizer} (ACVQ) that dynamically aligns code vectors with the data manifold to prevent collapse and improve quantization stability.

\textbf{3. We achieve state-of-the-art AF detection performance across diverse settings}—demonstrating strong results in standard intra-dataset 12-lead evaluation, challenging cross-dataset arbitrary-lead generalization, and realistic noise conditions, highlighting clinical applicability.

\section{Related Work}

Automatic detection of AF increasingly leveraged deep learning models in recent years. Early approaches employed CNNs \cite{cnn}\cite{fixed2} and RNNs \cite{rnn} to capture local patterns and long-range dependencies, respectively. Subsequent work introduced GNNs \cite{gnn} to model spatial correlations among multi-lead ECG signals, while more recent architectures—such as Transformers \cite{transformer} and Mamba \cite{mamba}—enabled efficient modeling of global context, steadily improving detection performance.
Existing methods largely fell into two paradigms: (1) fixed-lead training, which achieved strong performance on standard benchmarks but relied heavily on predefined lead configurations; and (2) random lead masking during training, which enhanced deployment flexibility but often compromised discriminative capability when critical leads were missing.

To mitigate information loss from limited leads, ECG lead reconstruction was explored as a preprocessing step, recovering a full 12-lead signal from a subset to provide richer input for downstream AF detection. Representative approaches included encoder-decoder CNNs/RNNs \cite{ed1}\cite{ed2}\cite{ed3}\cite{ed4}, GANs \cite{gan1}\cite{gan2} that improved waveform realism through adversarial training, VAEs \cite{vae} that modeled probabilistic latent representations, DDPMs \cite{ddpm} that generated high-fidelity signals via iterative refinement, and echo state networks (ESNs) \cite{esn} that captured complex temporal dynamics with recurrent reservoir structures. However, these methods primarily prioritized signal fidelity and remained largely decoupled from the AF detection task, lacking explicit optimization for diagnostically relevant features.

Even under full-lead conditions, AF detectors often suffered from severe performance degradation. This degradation stemmed from two coexisting challenges: (1) domain shift caused by variations in acquisition hardware and patient demographics; and (2) signal corruption due to environmental and physiological noise. To address these issues, prior work explored test-time adaptation \cite{general2} and domain generalization \cite{general1} to improve cross-domain generalization, while data augmentation \cite{robust} was employed to enhance robustness against noisy perturbations.
However, existing approaches typically treated signal reconstruction, AF detection, domain generalization, and noise robustness as largely decoupled objectives. As a result, they failed to leverage the synergies among these tasks within a coherent, unified architecture.

\section{Method}

\begin{figure*}
	\centering
	\includegraphics[width=1.0\textwidth]{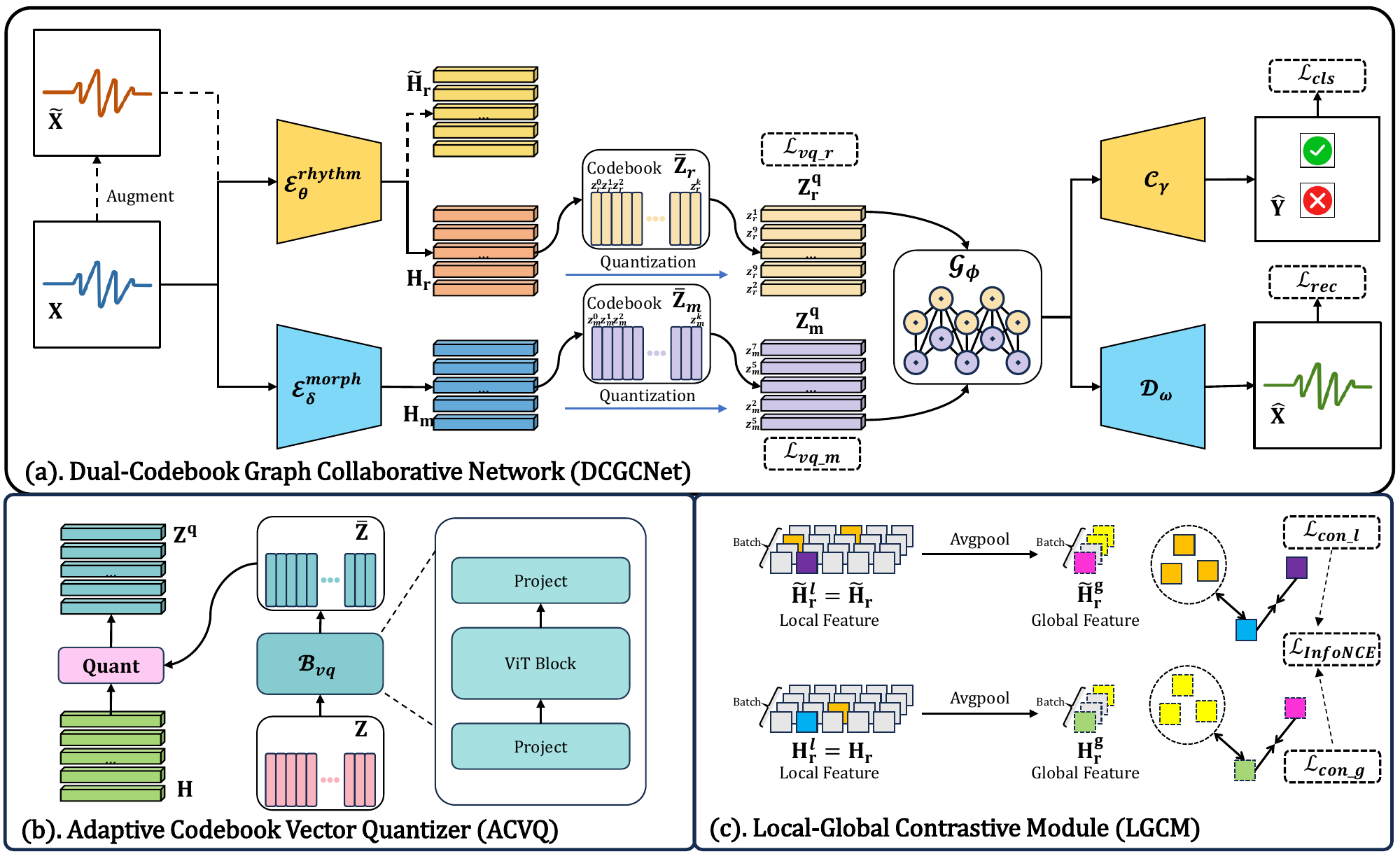}
	\caption{An end-to-end unified model integrating morphology-rhythm dual encoding, adaptive codebook quantization, local-global contrastive learning and heterogeneous graph interaction for multi-objective ECG analysis.}
	\label{figure:framework}
\end{figure*}

To simultaneously address AF detection and multi-lead ECG reconstruction, we propose the \textit{Dual-Codebook Graph Collaborative Network} (DCGCNet)—a unified framework that integrates four key components: morphology-rhythm dual encoding, adaptive codebook quantization, heterogeneous graph interaction, and local-global contrastive learning. As illustrated in Figure~\ref{figure:framework}(a), these components operate within a multi-task architecture where the two objectives mutually reinforce each other, enabling DCGCNet to learn structured and multi-scale representations that jointly enhance diagnostic accuracy and reconstruction fidelity.

\subsection{Rhythm-Morphology Dual Encoding}
AF is defined by disorganized atrial activation and an irregular ventricular response \cite{rhythm}. AF detection algorithms primarily rely on irregular RR intervals as a surrogate marker. Although the absence of organized P waves supports the diagnosis, it is often obscured by noise or fibrillatory activity. Consequently, AF detection depends primarily on \textbf{local rhythm} dynamics, especially beat-to-beat RR variability. In contrast, multi-lead ECG reconstruction requires faithful preservation of \textbf{global morphological} integrity, including the amplitude, duration, and spatiotemporal alignment of P-QRS-T complexes across leads—features critical for accurate electrophysiological interpretation and clinical decision-making \cite{morph}.

To explicitly account for the divergent signal characteristics underlying AF detection and ECG reconstruction, we propose two dedicated encoders with tailored receptive fields that extract rhythm- and morphology-specific representations from the input ECG $\mathbf{X} \in \mathbb{R}^{B \times C \times T}$ ($C$=12 leads, $T$=2500 sampling points, $B$ is batchsize):
\begin{align}
    \mathbf{H}_r &= \mathcal{E}_{\theta}^{rhythm}(\mathbf{X}) \in \mathbb{R}^{B \times T' \times D}, \\
    \mathbf{H}_m &= \mathcal{E}_{\delta}^{morph}(\mathbf{X}) \in \mathbb{R}^{B \times T' \times D},
\end{align}
where $T'$=$625$ and $D$=$256$. The rhythm encoder $\mathcal{E}_{\theta}^{rhythm}$ adopts convolutional small kernels ($k$=$5$) to model fine-grained irregularities characteristic of atrial fibrillation, whereas the morphology encoder $\mathcal{E}_{\delta}^{morph}$ utilizes larger kernels ($k$=$15$) to capture long-range structural dependencies essential for reconstruction.

\subsection{Adaptive Codebook Vector Quantizer}
The codebook in VQ-VAE provides a discrete latent representation that is inherently well-suited for generative modeling, making it highly effective for ECG reconstruction \cite{vqvae}\cite{vqvaeapp}. Moreover, by clustering semantically similar ECG segments into shared discrete codes, the codebook facilitates prototype-like learning that enhances feature consistency, improves generalization, and thereby enables robust AF detection even under distribution shifts \cite{dg}\cite{dgapp}.

To address the well-known issue of \textbf{codebook collapse} in vector quantization—where only a few codewords are actively used while others remain dormant \cite{collapse}—we propose the \textbf{ACVQ} as shown in Figure \ref{figure:framework}(b). Unlike conventional static codebooks, ACVQ generates context-aware codewords by modeling interactions among prototype vectors via a lightweight Transformer-based module, termed \textbf{VQBridge} $\mathcal{B}_{\text{vq}}$~\cite{vqbridge}. Specifically, given a static codebook $\mathbf{Z}_\ast \in \mathbb{R}^{K \times D}$ (where $\ast \in \{m, r\}$ denotes morphology or rhythm), the adaptive codebook $\overline{\mathbf{Z}}_\ast$ is computed as:
\begin{equation}
    \overline{\mathbf{Z}}_\ast = \mathbf{W}_{\text{out}} \cdot \mathrm{Transformer} \big( \mathbf{Z}_\ast \cdot \mathbf{W}_{\text{in}} \big), \quad \ast \in \{m, r\},
\end{equation}
where $\mathbf{W}_{\text{in}} \in \mathbb{R}^{D \times d'}$ and $\mathbf{W}_{\text{out}} \in \mathbb{R}^{d' \times D}$ are learnable projection matrices ($d' = 256$). This dynamic refinement encourages diverse codeword usage and mitigates dominance by a small subset of prototypes.

Then, we design a \textbf{dual-codebook VQ-VAE} that disentangles cardiac \textit{morphology} and \textit{rhythm} into two dedicated latent streams, leveraging the multi-scale features extracted by the hierarchical convolutional modules described in Section~3.1. The model employs two independent adaptive codebooks: a morphology codebook $\overline{\mathbf{Z}}_m$ and a rhythm codebook $\overline{\mathbf{Z}}_r$, each derived from its static counterpart via ACVQ. Given input features $\mathbf{H}_m$ and $\mathbf{H}_r$, quantization is performed separately through nearest-neighbor lookup:
\begin{equation}
    \mathbf{Z}_{\ast,i}^q = \arg\min_{k \in \{1,\dots,K\}} \big\| \mathbf{H}_{\ast,i} - \overline{\mathbf{Z}}_{\ast,k} \big\|_2^2,
    \quad \ast \in \{m, r\}.
\end{equation}
The standard vector quantization loss is applied independently to each stream:
\begin{equation}
    \mathcal{L}_{\text{vq}\_\ast} = \beta \big\| {\mathbf{Z}_{\ast}^{q}}^{\text{detach}} - \mathbf{H}_\ast \big\|_2^2 + \big\| \mathbf{Z}_{\ast}^{q} - \mathbf{H}_\ast^{\text{detach}} \big\|_2^2, \quad \ast \in \{m, r\},
\end{equation}
with gradients estimated via the straight-through estimator.

\subsection{Heterogeneous Graph Interaction}
Although rhythm and morphology serve different diagnostic purposes, they are interrelated: temporal changes in waveform shape often reflect rhythm disturbances, and rhythm irregularities can also appear as subtle morphological changes across beats.

To better integrate complementary rhythm and morphology cues in a temporally coherent manner, we construct a heterogeneous graph $\mathcal{G_{\phi}}$= $(\mathcal{V}, \mathcal{E})$ over the quantized token sequence of a single recording. The node set $\mathcal{V} = \{ r_t, m_t \}_{t=1}^{T'}$ contains rhythm and morphology tokens at each time step. Edges follow two principles as shown in Figure\ref{figure:framework} (a) $\mathcal{G_{\phi}}$: (i) \textit{intra-modality continuity}: each token connects to its immediate temporal neighbors within the same modality ($m_t \leftrightarrow m_{t\pm1}$, $r_t \leftrightarrow r_{t\pm1}$); (ii) \textit{inter-modality alignment}: each morphology token $m_t$ links to rhythm tokens at the same and adjacent positions ($r_{t-1}, r_t, r_{t+1}$), and symmetrically for $r_t$.

This design explicitly captures cross-modal interactions—both synchronous and temporally offset—within a three-step neighborhood, facilitating context-aware fusion of multi-modal temporal features for downstream tasks.

These structured relations are processed by a multi-head GAT \cite{gat} that computes contextualized node representations through adaptive neighbor aggregation, $\mathbf{x}_{i\_new} = \sum_{j \in \mathcal{N}(i)} \alpha_{ij} \, \mathbf{W} \mathbf{x}_j$, and:
\begin{equation}
    \alpha_{ij} = \frac{\exp\big(\mathrm{LeakyReLU}(\mathbf{a}^\top [\mathbf{W}\mathbf{x}_i \,\|\, \mathbf{W}\mathbf{x}_j])\big)}{\sum_{l \in \mathcal{N}(i)} \exp\big(\mathrm{LeakyReLU}(\mathbf{a}^\top [\mathbf{W}\mathbf{x}_i \,\|\, \mathbf{W}\mathbf{x}_l])\big)},
\end{equation}
where $\mathbf{x}_{i/j}$ denotes the input feature of node $i/j$, $\mathbf{W}$ is a learnable projection matrix, and $\alpha_{ij}$ represents the attention weight assigned to neighbor $j$ when updating node $i$.

\subsection{Local-Global Contrastive Module}
To improve robustness against common ECG artifacts, we adopt contrastive learning on rhythm features extracted from both the original view ($\mathbf{X}$) and an augmented view ($\tilde{\mathbf{X}}$) as described in Section~4.3. By enforcing consistency between ($\mathbf{H}$) and ($\tilde{\mathbf{H}}$) derived from differently perturbed versions of the same input, the model learns noise-invariant rhythm embeddings.

Specifically, we construct two complementary views of each sequence as shown in Figure~\ref{figure:framework}(c): a global representation capturing overall temporal semantics, and a local representation preserving fine-grained temporal structure:
\begin{align}
    &\quad \tilde{\mathbf{H}}_r^g/\mathbf{H}_r^g = \frac{1}{T'} \sum_{t=1}^{T'} \tilde{\mathbf{H}}_r/\mathbf{H}_r \in \mathbb{R}^D, \\
    &\quad \tilde{\mathbf{H}}_r^l/\mathbf{H}_r^l = \mathrm{flatten}\big(\{\tilde{\mathbf{H}}_r/\mathbf{H}_r\}_{t=1}^{T'}\big) \in \mathbb{R}^{T'D}.
\end{align}
For a mini-batch of $B$ samples, we compute both global ($\tilde{\mathbf{H}}_r^g,\mathbf{H}_r^g$) and local ($\tilde{\mathbf{H}}_r^l,\mathbf{H}_r^l$) representations for each sample, resulting in $2B$ embeddings.

We apply the symmetric InfoNCE loss \cite{contra} to align each original–augmented pair while repelling other (negative) samples in the batch:
\begin{equation}
    \mathcal{L}_{\text{InfoNCE}}(\mathbf{u}, \mathbf{v}) = -\log \frac{\exp(\mathbf{u}^\top \mathbf{v}/\tau)}{\sum_{\substack{(\mathbf{u}', \mathbf{v}') \in \mathcal{B} \\ (\mathbf{u}', \mathbf{v}') \neq (\mathbf{u}, \mathbf{v})}} \exp(\mathbf{u}^\top \mathbf{v}'/\tau)},
\end{equation}
where $(\mathbf{u}, \mathbf{v})$ denotes a positive pair, $(\tilde{\mathbf{H}}_r^g,\mathbf{H}_r^g)(\mathcal{L}_{\text{con}\_g})$ and $(\tilde{\mathbf{H}}_r^l,\mathbf{H}_r^l)(\mathcal{L}_{\text{con}\_l})$, and $\mathcal{B}$ is the set of all $2B$ embeddings in the current batch.

\subsection{Multi-Objective Optimization}
The total loss integrates six objectives that jointly address clinical requirements of AF detection, ECG reconstruction, domain generalization, and noise robustness:
\begin{equation}
\begin{split}
    \mathcal{L}_{\text{total}} ={}&
    \underbrace{\lambda_{\text{cls}} \mathcal{L}_{\text{cls}}}_{\text{AF Detection}} +
    \underbrace{\lambda_{\text{rec}} \mathcal{L}_{\text{rec}}}_{\text{ECG Reconstruction}} \\
    &+ \underbrace{
        \lambda_{\text{vq}} \mathcal{L}_{\text{vq\_r}} +
        \lambda_{\text{vq}} \mathcal{L}_{\text{vq\_m}}
    }_{\substack{\text{Quantitative Learning}}} +
    \underbrace{
        \lambda_{\text{cong}} \mathcal{L}_{\text{con\_g}} +
        \lambda_{\text{conl}} \mathcal{L}_{\text{con\_l}}
    }_{\substack{\text{Contrastive Learning}}},
\end{split}
\end{equation}
where $\mathcal{L}_{\text{cls}} = \text{CE}(\mathbf{Y}, \hat{\mathbf{Y}})$, $\mathcal{L}_{\text{rec}} = \|\mathbf{X} - \hat{\mathbf{X}}\|_2$.

\section{Experiments}
\subsection{Datasets}

\begin{figure*}
	\centering
	\includegraphics[width=0.7\textwidth]{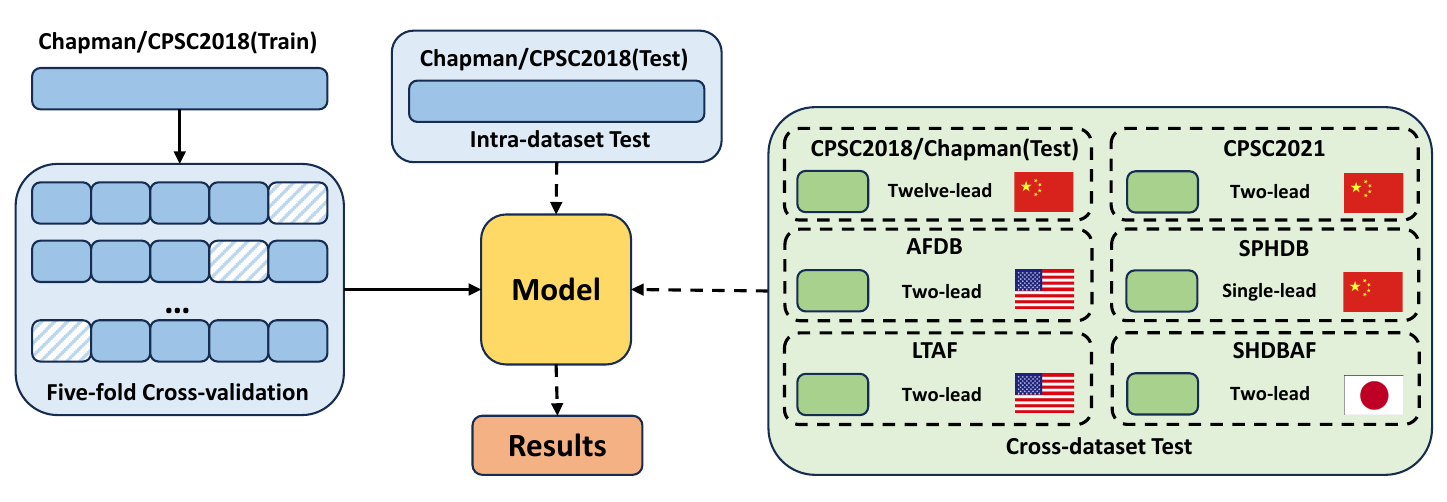}
	\caption{
        Experimental Protocol for Intra- and Cross-Dataset Evaluation of AF Detection Models.
        }
	\label{fig:trainprocess}
\end{figure*}

\begin{table}
\centering
\footnotesize
\setlength{\tabcolsep}{1.0pt}
\renewcommand{\arraystretch}{1.0}
\caption{Summary of the segmented datasets used in this study. The number of ``Normal'' and ``AF'' samples are reported after 10-second segmentation.}
\label{tab:datasetstats}
\begin{tabular}{l|ccccc}
\hline
Dataset & Normal & AF & Hz  & Country & Leads \\
\hline
Chapman (Train)   &6508        &1416        & 500 & China   & 12 \\
CPSC2018 (Train)  &1710        &2215        & 500 & China   & 12 \\
Chapman (Test)    &1617        &364        & 500 & China   & 12 \\
CPSC2018 (Test)   &423        &531        & 500 & China   & 12 \\
AFDB              &36718        &22053        & 250 & America & 2  \\
LTAF              &187470        &154590        & 128 & America & 2  \\
CPSC2021          &89570        &47440        & 200 & China   & 2  \\
SPHDB             &125000        &125000        & 200 & China   & 1  \\
SHDBAF           &614413        &148651        & 200 & Japan   & 2  \\
\hline
\end{tabular}
\end{table}

To comprehensively evaluate the performance and generalizability of our proposed model for arbitrary AF detection and ECG reconstruction, we conducted experiments on seven publicly available ECG databases. The \textbf{Chapman} and \textbf{CPSC2018} datasets were used for model development and internal validation via five-fold cross-validation. The remaining five datasets—\textbf{AFDB}, \textbf{LTAF}, \textbf{CPSC2021}, \textbf{SPHDB}, and \textbf{SHDBAF}—were reserved exclusively for external evaluation to assess the model’s generalizability across diverse populations, recording devices, and clinical settings, without any involvement in the training process. The complete experimental pipeline is illustrated in Figure~\ref{fig:trainprocess}.

Specifically, the \textbf{Chapman dataset \cite{chapman}} is a large-scale 12-lead ECG database established through a collaboration between Chapman University, Shaoxing People's Hospital, and Ningbo First Hospital. It comprises recordings from 45152 patients (500~Hz sampling rate, 10-second duration). All records are expert-annotated. For this study, we extracted a subset containing only ``Normal'' and ``AF'' labels, resulting in 9,905 segments.

The \textbf{CPSC2018 dataset \cite{cpsc2018}}, originating from the 2018 China Physiological Signal Challenge, contains 6877 12-lead ECG recordings (500~Hz, 6–60 seconds in length) collected from 11 hospitals. The data encompasses various cardiac arrhythmias. We selected all samples labeled as ``Normal'' or ``AF,'' which amounted to 4889 segments after being uniformly segmented into 10-second clips.

For external zero-shot evaluation—where the model parameters remain fixed—we processed the following five independent databases, extracting only ``Normal'' and ``AF'' segments and segmenting all records into 10-second windows:

    \textbf{AFDB (MIT-BIH Atrial Fibrillation Database) \cite{afdb}:} Contains 23 fully annotated 10-hour Holter recordings (250~Hz, 2 leads) from Beth Israel Deaconess Medical Center, with beat-wise rhythm labels including AFIB, AFL, J, and N.
    
    \textbf{LTAF (Long-Term AF Database) \cite{ltaf}:} Comprises 84 long-term ECG recordings (24–25 hours, 128~Hz, 2 leads) from subjects with paroxysmal or persistent AF. The heartbeat annotations (Normal/AF) were coordinated by MEDICALgorithmics and contributed to PhysioNet by Northwestern University.
    
    \textbf{CPSC2021 (the 2021 China Physiological Signal Challenge) \cite{cpsc2021}:} A dataset specifically designed for paroxysmal AF (PAF) detection, featuring Holter recordings (200~Hz, 2 leads) with precise onset/offset timestamps for AF episodes and three primary labels: AF, PAF, and Normal.
    
    \textbf{SPHDB (Shandong Provincial Hospital Database) \cite{sphdb}:} Includes 24-hour wearable ECG recordings from 250 PAF patients (200~Hz, 1 lead), with all heartbeats clinically annotated as AF or non-AF.
    
    \textbf{SHDBAF (Saitama Heart Database Atrial Fibrillation) \cite{shdbaf}:} A Japanese open-source Holter database containing 128 24-hour recordings (200~Hz, 2 leads: modified CC5 and NASA). Of these, 98 recordings from 93 subjects have expert cardiologist-provided beat-wise annotations (AFIB, AFL, AT, PAT, N).

 All ECG segments were resampled to a uniform sampling rate of 250~Hz prior to train and evaluation. The final statistics of the segmented datasets used in our experiments are summarized in Table~\ref{tab:datasetstats}.

\subsection{Evaluation Metrics}
We adopt distinct evaluation protocols for the two primary tasks. For the \textbf{ECG lead reconstruction task}, performance is measured using four signal fidelity metrics: the \textbf{Pearson Correlation Coefficient (PCC)} captures linear waveform similarity; the \textbf{Coefficient of Determination ($R^2$)} reflects the proportion of variance in the ground-truth signal explained by the reconstruction; the \textbf{Root Mean Squared Error (RMSE)} emphasizes large deviations through quadratic weighting; and the \textbf{Mean Absolute Error (MAE)} provides an interpretable average error magnitude in microvolts ($\mu$V).  

For the \textbf{AF detection task}, we report five standard segment-level classification metrics: the \textbf{Area Under the ROC Curve (AUC)} assesses threshold-invariant discriminative ability; \textbf{Accuracy (ACC)} indicates overall correctness; the \textbf{F1-score (F1)} balances precision and recall via their harmonic mean; \textbf{Precision (Pre)} quantifies the reliability of positive predictions; and \textbf{Recall} measures sensitivity to true AF episodes.

\begin{table*}[t]
\centering
\footnotesize
\setlength{\tabcolsep}{1.0pt}
\renewcommand{\arraystretch}{1.0}
\caption{Performance Comparison of Different Models for Lead Reconstruction Performance on Chapman and CPSC2018 Datasets (mean $\pm$ std).
$\uparrow$: higher is better; $\downarrow$: lower is better.
Bold: best result; underlined: second-best result.
}
\label{tab:reconstruction}
\begin{tabular}{l|cccc|cccc|cccc|cccc}
\hline
\multirow{2}{*}{Model} & 
\multicolumn{4}{c|}{Chapman Lead I} & 
\multicolumn{4}{c|}{Chapman Lead I+II+V1} & 
\multicolumn{4}{c|}{CPSC2018 Lead I} & 
\multicolumn{4}{c}{CPSC2018 Lead I+II+V1} \\
\cline{2-17}
& PCC$\uparrow$ & R$^2$$\uparrow$ & RMSE$\downarrow$ & MAE$\downarrow$ &
  PCC$\uparrow$ & R$^2$$\uparrow$ & RMSE$\downarrow$ & MAE$\downarrow$ &
  PCC$\uparrow$ & R$^2$$\uparrow$ & RMSE$\downarrow$ & MAE$\downarrow$ &
  PCC$\uparrow$ & R$^2$$\uparrow$ & RMSE$\downarrow$ & MAE$\downarrow$ \\
\hline
ResCNN\cite{ed1} &
\makecell[t]{\underline{0.7828} \\ {\tiny(0.0012)}} &
\makecell[t]{\textbf{0.6424} \\ {\tiny(0.0016)}} &
\makecell[t]{\underline{0.5983} \\ {\tiny(0.0014)}} &
\makecell[t]{\underline{0.3173} \\ {\tiny(0.0012)}} &
\makecell[t]{\underline{0.9349} \\ {\tiny(0.0006)}} &
\makecell[t]{\textbf{0.8881} \\ {\tiny(0.0012)}} &
\makecell[t]{\textbf{0.3402} \\ {\tiny(0.0020)}} &
\makecell[t]{\underline{0.1643} \\ {\tiny(0.0016)}} &
\makecell[t]{\underline{0.7519} \\ {\tiny(0.0013)}} &
\makecell[t]{0.5933 \\ {\tiny(0.0053)}} &
\makecell[t]{0.6406 \\ {\tiny(0.0042)}} &
\makecell[t]{\underline{0.3387} \\ {\tiny(0.0016)}} &
\makecell[t]{0.9132 \\ {\tiny(0.0011)}} &
\makecell[t]{\textbf{0.8537} \\ {\tiny(0.0021)}} &
\makecell[t]{\textbf{0.3853} \\ {\tiny(0.0029)}} &
\makecell[t]{\underline{0.1800} \\ {\tiny(0.0022)}} \\

GRU\cite{ed2} &
\makecell[t]{0.7770 \\ {\tiny(0.0010)}} &
\makecell[t]{0.6328 \\ {\tiny(0.0018)}} &
\makecell[t]{0.6064 \\ {\tiny(0.0015)}} &
\makecell[t]{0.3257 \\ {\tiny(0.0023)}} &
\makecell[t]{0.9336 \\ {\tiny(0.0011)}} &
\makecell[t]{\underline{0.8844} \\ {\tiny(0.0021)}} &
\makecell[t]{0.3460 \\ {\tiny(0.0033)}} &
\makecell[t]{0.1768 \\ {\tiny(0.0021)}} &
\makecell[t]{0.7541 \\ {\tiny(0.0047)}} &
\makecell[t]{\textbf{0.6025} \\ {\tiny(0.0048)}} &
\makecell[t]{\underline{0.6332} \\ {\tiny(0.0038)}} &
\makecell[t]{0.3397 \\ {\tiny(0.0025)}} &
\makecell[t]{0.9115 \\ {\tiny(0.0006)}} &
\makecell[t]{\underline{0.8521} \\ {\tiny(0.0014)}} &
\makecell[t]{\underline{0.3875} \\ {\tiny(0.0019)}} &
\makecell[t]{0.1910 \\ {\tiny(0.0082)}} \\

EKGAN\cite{gan2} &
\makecell[t]{0.7272 \\ {\tiny(0.0011)}} &
\makecell[t]{0.5342 \\ {\tiny(0.0122)}} &
\makecell[t]{0.6801 \\ {\tiny(0.0089)}} &
\makecell[t]{0.3714 \\ {\tiny(0.0043)}} &
\makecell[t]{0.9139 \\ {\tiny(0.0016)}} &
\makecell[t]{0.8320 \\ {\tiny(0.0042)}} &
\makecell[t]{0.4084 \\ {\tiny(0.0050)}} &
\makecell[t]{0.1983 \\ {\tiny(0.0047)}} &
\makecell[t]{0.7309 \\ {\tiny(0.0025)}} &
\makecell[t]{0.5177 \\ {\tiny(0.0202)}} &
\makecell[t]{0.6941 \\ {\tiny(0.0144)}} &
\makecell[t]{0.3652 \\ {\tiny(0.0048)}} &
\makecell[t]{\textbf{0.9190} \\ {\tiny(0.0026)}} &
\makecell[t]{0.8441 \\ {\tiny(0.0046)}} &
\makecell[t]{0.3947 \\ {\tiny(0.0058)}} &
\makecell[t]{0.1826 \\ {\tiny(0.0043)}} \\

ESN\cite{esn} &
\makecell[t]{0.6977 \\ {\tiny(0.0003)}} &
\makecell[t]{0.5276 \\ {\tiny(0.0014)}} &
\makecell[t]{0.6849 \\ {\tiny(0.0010)}} &
\makecell[t]{0.3624 \\ {\tiny(0.0002)}} &
\makecell[t]{0.9099 \\ {\tiny(0.0002)}} &
\makecell[t]{0.8261 \\ {\tiny(0.0003)}} &
\makecell[t]{0.4156 \\ {\tiny(0.0004)}} &
\makecell[t]{0.1921 \\ {\tiny(0.0003)}} &
\makecell[t]{0.6617 \\ {\tiny(0.0012)}} &
\makecell[t]{0.4825 \\ {\tiny(0.0010)}} &
\makecell[t]{0.7191 \\ {\tiny(0.0007)}} &
\makecell[t]{0.3754 \\ {\tiny(0.0006)}} &
\makecell[t]{0.8893 \\ {\tiny(0.0004)}} &
\makecell[t]{0.7997 \\ {\tiny(0.0014)}} &
\makecell[t]{0.4474 \\ {\tiny(0.0015)}} &
\makecell[t]{0.1988 \\ {\tiny(0.0010)}} \\
\hline
DCGCNet(Ours) &
\makecell[t]{\textbf{0.7865} \\ {\tiny(0.0015)}} &
\makecell[t]{\underline{0.6407} \\ {\tiny(0.0025)}} &
\makecell[t]{\textbf{0.5974} \\ {\tiny(0.0021)}} &
\makecell[t]{\textbf{0.3086} \\ {\tiny(0.0009)}} &
\makecell[t]{\textbf{0.9376} \\ {\tiny(0.0009)}} &
\makecell[t]{0.8811 \\ {\tiny(0.0019)}} &
\makecell[t]{\underline{0.3437} \\ {\tiny(0.0027)}} &
\makecell[t]{\textbf{0.1530} \\ {\tiny(0.0018)}} &
\makecell[t]{\textbf{0.7583} \\ {\tiny(0.0015)}} &
\makecell[t]{\underline{0.6012} \\ {\tiny(0.0058)}} &
\makecell[t]{\textbf{0.6313} \\ {\tiny(0.0046)}} &
\makecell[t]{\textbf{0.3259} \\ {\tiny(0.0014)}} &
\makecell[t]{\underline{0.9158} \\ {\tiny(0.0010)}} &
\makecell[t]{0.8448 \\ {\tiny(0.0021)}} &
\makecell[t]{0.3938 \\ {\tiny(0.0027)}} &
\makecell[t]{\textbf{0.1714} \\ {\tiny(0.0011)}} \\
\hline
\end{tabular}
\end{table*}

\begin{table*}[t]
\centering
\footnotesize
\setlength{\tabcolsep}{8pt}
\renewcommand{\arraystretch}{1.0}
\caption{Performance Comparison of Different Models for Twelve-Lead Atrial Fibrillation Detection on Chapman and CPSC2018 Datasets (mean $\pm$ std). Bold: best result}
\label{tab:performance}
\begin{tabular}{l|ccccc|ccccc}
\hline
\multirow{2}{*}{Model} & 
\multicolumn{5}{c|}{Chapman Dataset} & 
\multicolumn{5}{c}{CPSC2018 Dataset} \\
\cline{2-11}
& AUC & ACC & F1 & Pre & Recall & AUC & ACC & F1 & Pre & Recall \\
\hline
DNN\cite{fixed2} &
\makecell[t]{0.9790 \\ {\tiny(0.0027)}} &
\makecell[t]{0.9332 \\ {\tiny(0.0033)}} &
\makecell[t]{0.8840 \\ {\tiny(0.0053)}} &
\makecell[t]{0.9016 \\ {\tiny(0.0097)}} &
\makecell[t]{0.8692 \\ {\tiny(0.0080)}} &
\makecell[t]{0.9841 \\ {\tiny(0.0019)}} &
\makecell[t]{0.9306 \\ {\tiny(0.0099)}} &
\makecell[t]{0.9301 \\ {\tiny(0.0097)}} &
\makecell[t]{0.9293 \\ {\tiny(0.0090)}} &
\makecell[t]{0.9330 \\ {\tiny(0.0082)}} \\

DenseRNN\cite{rnn} &
\makecell[t]{0.9974 \\ {\tiny(0.0003)}} &
\makecell[t]{0.9694 \\ {\tiny(0.0047)}} &
\makecell[t]{0.9459 \\ {\tiny(0.0089)}} &
\makecell[t]{0.9759 \\ {\tiny(0.0034)}} &
\makecell[t]{0.9217 \\ {\tiny(0.0137)}} &
\makecell[t]{0.9965 \\ {\tiny(0.0005)}} &
\makecell[t]{0.9157 \\ {\tiny(0.0201)}} &
\makecell[t]{0.9156 \\ {\tiny(0.0201)}} &
\makecell[t]{0.9198 \\ {\tiny(0.0158)}} &
\makecell[t]{0.9238 \\ {\tiny(0.0181)}} \\

ConvRGNN\cite{gnn} &
\makecell[t]{0.9627 \\ {\tiny(0.0048)}} &
\makecell[t]{0.9318 \\ {\tiny(0.0040)}} &
\makecell[t]{0.8832 \\ {\tiny(0.0095)}} &
\makecell[t]{0.8948 \\ {\tiny(0.0086)}} &
\makecell[t]{0.8739 \\ {\tiny(0.0201)}} &
\makecell[t]{0.9545 \\ {\tiny(0.0091)}} &
\makecell[t]{0.8889 \\ {\tiny(0.0109)}} &
\makecell[t]{0.8875 \\ {\tiny(0.0112)}} &
\makecell[t]{0.8878 \\ {\tiny(0.0113)}} &
\makecell[t]{0.8881 \\ {\tiny(0.0120)}} \\

MSGformer\cite{transformer} &
\makecell[t]{0.9863 \\ {\tiny(0.0033)}} &
\makecell[t]{0.9556 \\ {\tiny(0.0052)}} &
\makecell[t]{0.9246 \\ {\tiny(0.0102)}} &
\makecell[t]{0.9322 \\ {\tiny(0.0077)}} &
\makecell[t]{0.9183 \\ {\tiny(0.0194)}} &
\makecell[t]{0.9962 \\ {\tiny(0.0004)}} &
\makecell[t]{0.9730 \\ {\tiny(0.0044)}} &
\makecell[t]{0.9727 \\ {\tiny(0.0043)}} &
\makecell[t]{0.9721 \\ {\tiny(0.0049)}} &
\makecell[t]{0.9735 \\ {\tiny(0.0036)}} \\

ECGMamba\cite{mamba} &
\makecell[t]{0.9910 \\ {\tiny(0.0012)}} &
\makecell[t]{0.9637 \\ {\tiny(0.0018)}} &
\makecell[t]{0.9385 \\ {\tiny(0.0034)}} &
\makecell[t]{0.9458 \\ {\tiny(0.0055)}} &
\makecell[t]{0.9317 \\ {\tiny(0.0076)}} &
\makecell[t]{0.9931 \\ {\tiny(0.0014)}} &
\makecell[t]{0.9602 \\ {\tiny(0.0036)}} &
\makecell[t]{0.9597 \\ {\tiny(0.0036)}} &
\makecell[t]{0.9595 \\ {\tiny(0.0041)}} &
\makecell[t]{0.9607 \\ {\tiny(0.0039)}} \\

RawECGNet\cite{general1} &
\makecell[t]{0.9951 \\ {\tiny(0.0027)}} &
\makecell[t]{0.9746 \\ {\tiny(0.0074)}} &
\makecell[t]{0.9583 \\ {\tiny(0.0126)}} &
\makecell[t]{0.9506 \\ {\tiny(0.0107)}} &
\makecell[t]{0.9670 \\ {\tiny(0.0187)}} &
\makecell[t]{0.9970 \\ {\tiny(0.0012)}} &
\makecell[t]{0.9819 \\ {\tiny(0.0041)}} &
\makecell[t]{0.9817 \\ {\tiny(0.0041)}} &
\makecell[t]{0.9817 \\ {\tiny(0.0042)}} &
\makecell[t]{0.9819 \\ {\tiny(0.0043)}} \\

TTDADualNet\cite{general2} &
\makecell[t]{0.9946 \\ {\tiny(0.0019)}} &
\makecell[t]{0.9763 \\ {\tiny(0.0025)}} &
\makecell[t]{0.9611 \\ {\tiny(0.0042)}} &
\makecell[t]{0.9535 \\ {\tiny(0.0043)}} &
\makecell[t]{0.9693 \\ {\tiny(0.0071)}} &
\makecell[t]{0.9974 \\ {\tiny(0.0019)}} &
\makecell[t]{0.9811 \\ {\tiny(0.0060)}} &
\makecell[t]{0.9809 \\ {\tiny(0.0060)}} &
\makecell[t]{0.9804 \\ {\tiny(0.0062)}} &
\makecell[t]{0.9816 \\ {\tiny(0.0058)}} \\
\hline
DCGCNet(Ours) &
\makecell[t]{\textbf{0.9998} \\ {\tiny(0.0001)}} &
\makecell[t]{\textbf{0.9966} \\ {\tiny(0.0013)}} &
\makecell[t]{\textbf{0.9943} \\ {\tiny(0.0021)}} &
\makecell[t]{\textbf{0.9924} \\ {\tiny(0.0032)}} &
\makecell[t]{\textbf{0.9962} \\ {\tiny(0.0015)}} &
\makecell[t]{\textbf{0.9991} \\ {\tiny(0.0005)}} &
\makecell[t]{\textbf{0.9937} \\ {\tiny(0.0017)}} &
\makecell[t]{\textbf{0.9936} \\ {\tiny(0.0017)}} &
\makecell[t]{\textbf{0.9932} \\ {\tiny(0.0016)}} &
\makecell[t]{\textbf{0.9941} \\ {\tiny(0.0018)}} \\
\hline
\end{tabular}
\end{table*}

\subsection{Implementation Details}
All experiments were implemented in PyTorch 2.4.1 with CUDA 11.5 acceleration and conducted on a workstation equipped with an NVIDIA GeForce RTX 3090 GPU (24 GB VRAM) and an Intel Core i7-12700KF CPU. The model architecture in Section~III employed a codebook size of $K = 512$ and a vector quantization loss weight $\beta = 0.25$. The composite loss function was defined as $\mathcal{L} = \lambda_{\text{cls}} \mathcal{L}_{\text{cls}} + \lambda_{\text{rec}} \mathcal{L}_{\text{rec}} + \lambda_{\text{vq}} \mathcal{L}_{\text{vq}} + \lambda_{\text{cong}} \mathcal{L}_{\text{cong}} + \lambda_{\text{conl}} \mathcal{L}_{\text{conl}}$, with weights $\lambda_{\text{cls}} = 1$, $\lambda_{\text{rec}} = 1$, $\lambda_{\text{vq}} = 0.2$, $\lambda_{\text{cong}} = 0.5$, and $\lambda_{\text{conl}} = 0.1$. A Gumbel-Softmax temperature of $\tau = 0.1$ was used for discrete token sampling.

In our contrastive learning setup, we applied two ECG-specific data augmentation strategies to generate positive pairs for training: (i) additive Gaussian noise with standard deviation sampled uniformly from $[0.1, 0.8]$ relative to the signal amplitude; and (ii) baseline wander with amplitude in $[0.1, 0.8]$, and frequency in $[0.05, 0.3]$ Hz. The model was optimized using the Adam optimizer with a batch size of 16, an initial learning rate of $1 \times 10^{-4}$, weight decay of $1 \times 10^{-4}$, and trained for 50 epochs without learning rate scheduling.

\begin{figure*}
	\centering
	\includegraphics[width=1.0\textwidth]{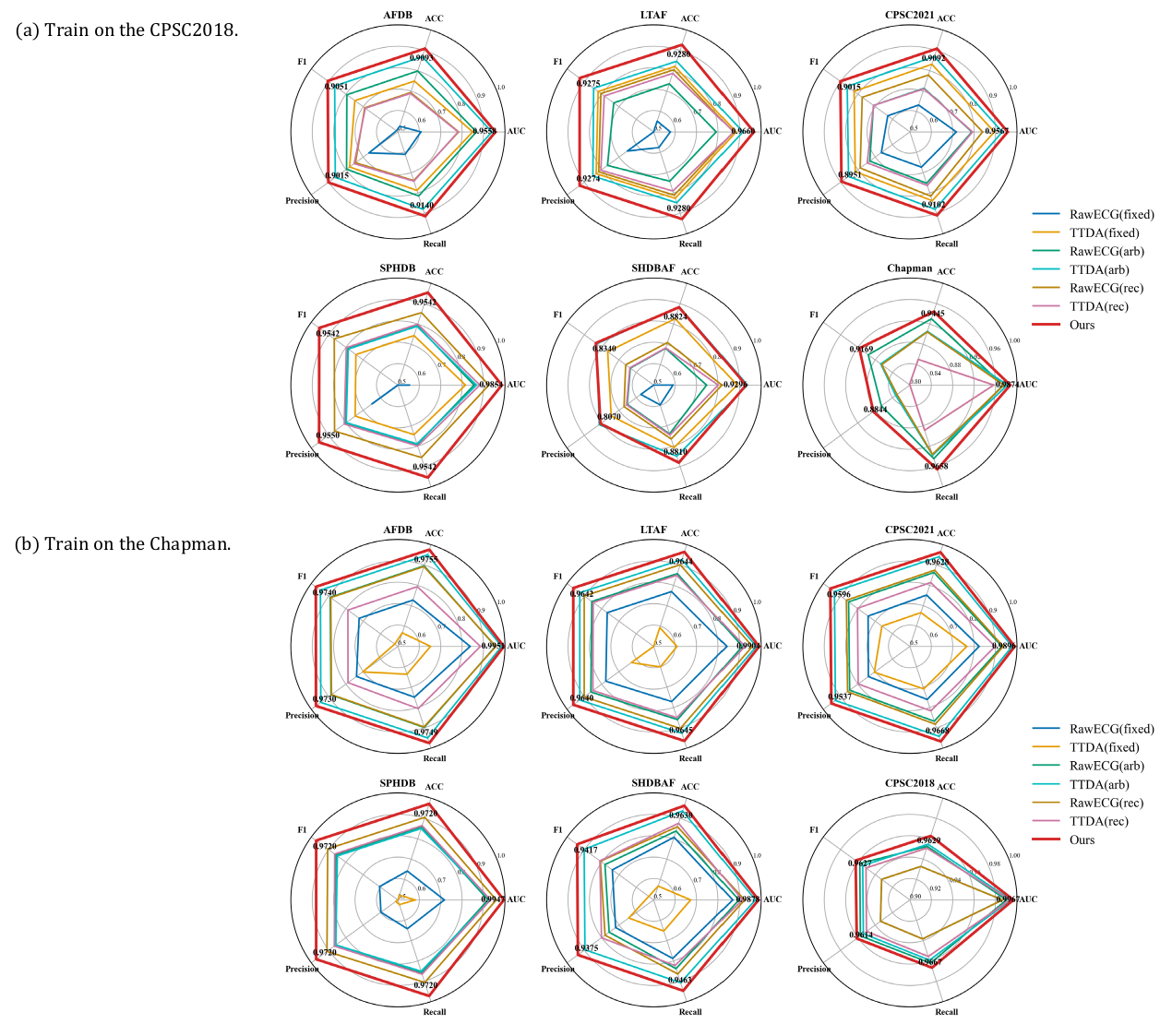}
	\caption{
        Cross-dataset generalization performance radar plots on AF detection comparing DCGCNet against baseline methods under three input strategies (fixed, arbitrary, reconstruction). (a) Models trained on CPSC2018 and evaluated on six external test sets. (b) Models trained on Chapman and tested on six datasets. Metrics shown include AUC, Accuracy, F1-score, Precision, and Recall. 
        }
	\label{fig:generalization}
\end{figure*}
\subsection{Comparative Experiments}
We compared our proposed DCGCNet against a diverse set of state-of-the-art baselines on two core tasks: ECG lead reconstruction and twelve-lead AF detection. To ensure a fair comparison, we reproduced all baseline methods under the same experimental setup. 

For the reconstruction task, we evaluated all models on the Chapman and CPSC2018 datasets under two input configurations—single-lead (Lead I) and three-lead (Leads I+II+V1)—using four fidelity metrics: PCC, $R^2$, RMSE, and MAE. Baseline methods include \textbf{ResCNN}\cite{ed1} (Original task: reconstructed 12-lead ECG from leads I, II, V3 using residual CNNs), \textbf{GRU}\cite{ed2} (Original task: from I and II via GRU), \textbf{EKGAN}\cite{gan2} (Original task: from lead I via GAN), and \textbf{ESN}\cite{esn} (Original task: from I and V1 using echo state networks). As shown in Table~\ref{tab:reconstruction}, DCGCNet consistently achieved the best or second-best performance across majority settings. Although ResCNN yielded slightly higher $R^2$ on Chapman Lead I and CPSC2018 Lead I+II+V1, DCGCNet exhibited more balanced improvements across all metrics. A reconstruction result was shown in Figure \ref{fig:reconstruction}

For AF detection, we benchmarked DCGCNet against seven recent deep learning architectures on the same datasets, reporting five standard classification metrics at the segment level. \textbf{DNN}\cite{fixed2} employed a standard deep neural network operating directly on 12-lead ECG inputs. \textbf{DenseRNN}\cite{rnn} integrated bidirectional recurrent units into a densely connected residual architecture to model temporal dynamics across all leads. \textbf{ConvRGNN}\cite{gnn} formulated the 12-lead ECG as a graph and leveraged convolutional residual graph neural networks to capture inter-lead spatial dependencies. \textbf{MSGformer}\cite{transformer} adopted a multi-scale grid attention mechanism to extract hierarchical temporal features from 12-lead signals. \textbf{ECGMamba}\cite{mamba} utilized a bidirectional state space model for efficient sequence modeling and classification. Additionally, two recent generalizable approaches were included: \textbf{RawECGNet}\cite{general1}, which incorporated batch normalization and an uncertainty-aware domain-shift module to improve robustness, and \textbf{TTDADualNet}\cite{general2}, which addressed cross-dataset generalization through test-time data augmentation. 

As summarized in Table~\ref{tab:performance}, DCGCNet established a new state of the art. On the Chapman dataset, it achieved an AUC of 0.9998 and an accuracy of 0.9966, with an F1-score that surpassed the strongest baseline (TTDADualNet) by 3.32 percentage points. On CPSC2018, DCGCNet improved accuracy by 1.26\% and F1-score by 1.27\% over TTDADualNet, while maintaining exceptional precision-recall balance. These results highlighted the effectiveness of our dual-path architecture and contextual graph-based representation in jointly capturing local rhythm dynamics and global inter-lead morphological dependencies. In addition, we gave the comparison of the results of the original literature in the Table\ref{tab:sota_comparison}.

\begin{figure*}
	\centering
	\includegraphics[width=1.0\textwidth]{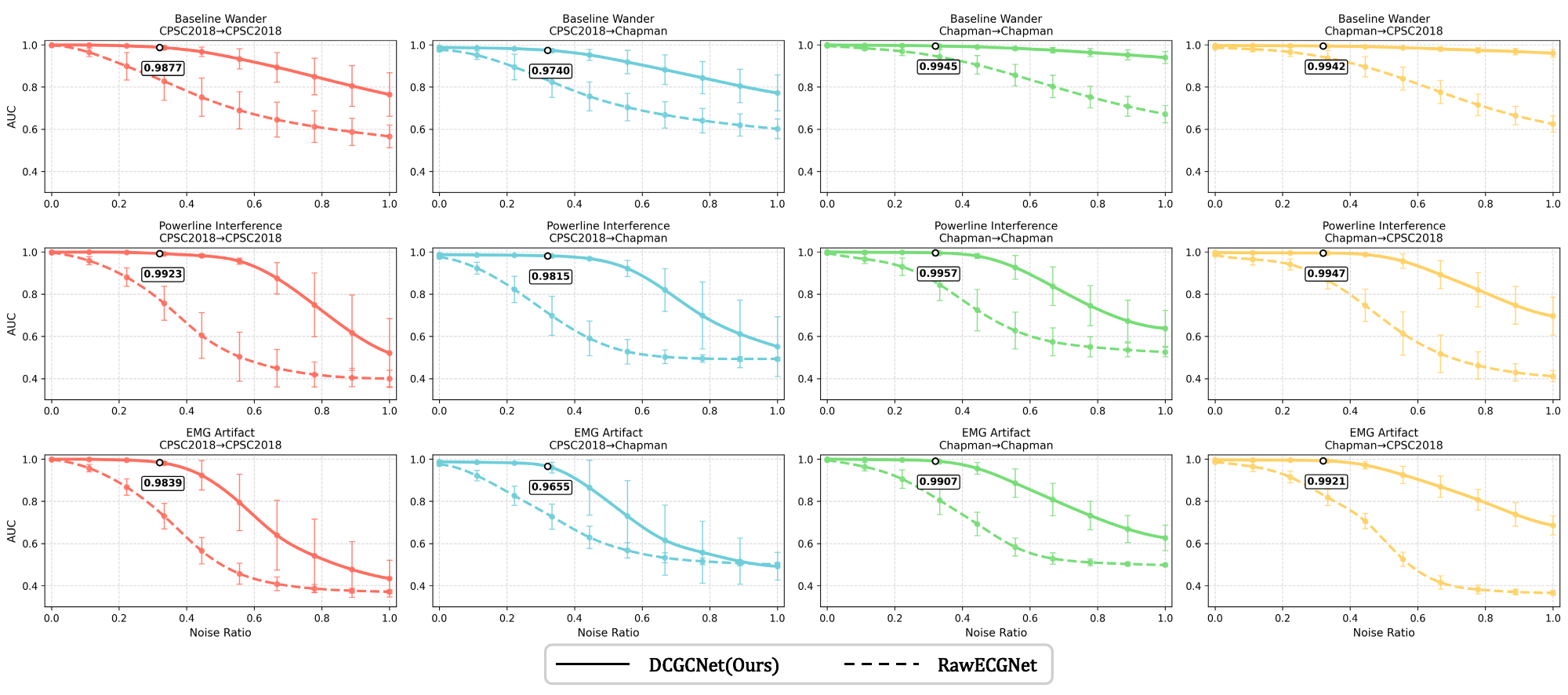}
	\caption{
        Robustness of DCGCNet and RawECGNet on AF detection under increasing levels of ECG artifacts. 
AUC performance is plotted across four training--testing configurations as a function of noise intensity, defined as the ratio of artifact amplitude to original ECG amplitude (Ratio). 
Each curve shows the mean AUC over five independent runs, with error bars indicating the corresponding standard deviation. 
The clinically relevant 10~dB SNR level ($\textit{Ratio} \approx 0.32$)—representative of typical ECG interference in wearable settings—is highlighted by white markers. 
        }
	\label{fig:robust}
\end{figure*}
\subsection{Generalization Experiments}
To rigorously evaluate the cross-dataset generalization capability of the proposed DCGCNet, we conducted comprehensive experiments under two distinct training regimes: (1) models trained on \textit{Chapman} and evaluated on six external test sets (AFDB, LTAF, CPSC2021, SPHDB, SHDBAF, and CPSC2018); and (2) models trained on \textit{CPSC2018} and tested on the same five datasets and Chapman. This design enabled a thorough assessment of model robustness across diverse acquisition protocols, patient populations, and labeling criteria.

For each generalization baseline method, RawECGNet\cite{general1} and TTDA-DualNet\cite{general2}, we implemented three variants corresponding to different input handling strategies: \textbf{Fixed}: trained on standard 12-lead ECGs and directly applied to test data without adaptation;
\textbf{Arbitrary}: trained on synthetically masked 12-lead ECGs (simulating variable lead availability) to learn from arbitrary lead configurations;
\textbf{Reconstruction}: trained on full 12-lead ECGs but, during inference, first reconstructed missing leads using a ResCNN\cite{ed1} before classification.

In contrast, our DCGCNet jointly optimized lead reconstruction and AF detection in an end-to-end manner, explicitly designed for arbitrary-lead inputs. All methods were reimplemented under identical experimental conditions—including data preprocessing, optimizer settings, and evaluation metrics—to ensure a fair comparison. 

As visualized in Fig.~\ref{fig:generalization}, DCGCNet consistently achieved state-of-the-art performance across all test sets. Notably, when trained on \textit{Chapman} and tested on \textit{CPSC2018} (both 12-lead datasets), the \textit{fixed} and \textit{reconstruction} variants yielded identical results, as no lead imputation was required—a pattern symmetrically observed in the reverse setting (\textit{CPSC2018} $\rightarrow$ \textit{Chapman}). In more challenging cross-domain scenarios involving variable lead availability (e.g., AFDB, SPHDB), the \textit{arbitrary} and \textit{reconstruction} strategies substantially improved over the \textit{fixed} baseline, yet still lagged behind DCGCNet by a clear margin in AUC and F1-score. These results demonstrated that the learnable codebook served as a unified representation bridge between reconstruction and classification, and its prototypical nature enabled it to generalize effectively across diverse clinical datasets.

\subsection{Robust Experiments}
To rigorously assess the robustness of DCGCNet under realistic clinical noise conditions, we conducted comprehensive experiments by progressively corrupting clean ECG segments with three dominant artifact types: baseline wander (BLW), powerline interference (PLI), and electromyographic (EMG) artifact. The corruption intensity was controlled by an SNR-equivalent parameter defined as the ratio of artifact amplitude to the original ECG amplitude (denoted as \textit{Ratio}). Following established literature\cite{emg}, which reported an average EMG artifact intensity of about 10 dB SNR ( $ \textit{Ratio} \approx 0.32 $ ), we adopted this level as a clinically relevant reference for ambulatory ECG. Although BLW and PLI were not similarly quantified, we used the same 10 dB SNR for all three artifacts to ensure consistent evaluation. A specific example of adding noise was shown in Figure \ref{fig:noise}.

Figure~\ref{fig:robust} showed the AUC performance of DCGCNet across four cross-dataset configurations as \textit{Ratio} increases from 0 to 1. Each curve represented the mean AUC over five independent runs, with error bars indicating the standard deviation. The results clearly demonstrated that DCGCNet consistently outperformed RawECGNet under all noise conditions. The 10~dB operating point ($\textit{Ratio}$=$0.32$) was explicitly marked by white circular markers. The results demonstrated that DCGCNet maintained consistently high performance even under severe noise corruption, with only marginal degradation at the clinically relevant 10~dB level—underscoring its robustness and practical applicability in real-world ECG monitoring scenarios.

\subsection{Ablation Experiments}
We conducted ablation studies to assess the contribution of each component in DCGCNet across three settings: intra-dataset performance, cross-dataset generalization, and robustness to ECG noise.

Intra-dataset performance (Table \ref{tab:ablation1}) showed that the reconstruction (rec) and classification (cls) tasks were mutually beneficial—each task enhances the learning of features relevant to the other. This also demonstrated that removing either the rhythm or morphology encoder degraded AF detection and signal reconstruction, respectively, highlighting their complementary roles. The Heterogeneous Graph Interaction (HGI), Adaptive Codebook Vector Quantizer (ACVQ) and Local-Global Contrastive Module (LGCM) also contributed notably to reconstruction fidelity and classification accuracy. Cross-dataset generalization (Table \ref{tab:ablation2}) revealed that discrete representation learning was critical for domain transfer. Removing the ACVQ caused substantially larger AUC drops than ablating the basic Adaptive Codebook (AC), confirming ACVQ’s importance in bridging domain gaps. Robustness to noise (Table \ref{tab:ablation3}) demonstrated that both ACVQ and LGCM were essential under common artifacts—BLW, PLI, and EMG.

Together, these results validated that the integration of Rhythm-Morphology Dual Encoding, ACVQ, LGCM, and HGI was key to DCGCNet’s strong and reliable AF detection performance. In addition, the sensitivity analysis of the loss function was shown in the Table\ref{tab:sensitivity}.

\begin{table}[h]
\footnotesize
\setlength{\tabcolsep}{3pt}
\centering
\caption{Ablation study on intra-dataset: Lead I ECG reconstruction (PCC) and AF detection (AUC).}
\label{tab:ablation1}
\begin{tabular}{lcccc}
\toprule
\multirow{2}{*}{Setting} & \multicolumn{2}{c}{Rec. (PCC)} & \multicolumn{2}{c}{Cls. (AUC)} \\
\cmidrule(lr){2-3} \cmidrule(lr){4-5}
& Chapman & CPSC2018 & Chapman & CPSC2018 \\
\midrule
DCGCNet          & 0.7865 & 0.7583 & 0.9998 & 0.9991 \\
w/o rec          & - & - & 0.9883 & 0.9749 \\
w/o cls          & 0.7751 & 0.7553 & - & - \\
w/o rhythm       & 0.7812 & 0.7521 & 0.9823 & 0.9754 \\
w/o morph        & 0.7205 & 0.6957 & 0.9986 & 0.9972 \\
w/o AC           & 0.7398 & 0.7092 & 0.9865 & 0.9798 \\
w/o ACVQ         & 0.7347 & 0.7043 & 0.9852 & 0.9784 \\
w/o LGCM\_L      & 0.7768 & 0.7425 & 0.9948 & 0.9897 \\
w/o LGCM\_G      & 0.7789 & 0.7462 & 0.9959 & 0.9915 \\
w/o LGCM         & 0.7724 & 0.7438 & 0.9927 & 0.9873 \\
w/o HGI         & 0.7742 & 0.7487 & 0.9863 & 0.9804 \\
\bottomrule
\end{tabular}
\end{table}

\begin{table}[h]
\footnotesize
\setlength{\tabcolsep}{1pt}
\centering
\caption{Ablation study on cross-dataset generalization for AF detection (trained on Chapman).}
\label{tab:ablation2}
\begin{tabular}{lccc}
\toprule
Setting & CPSC2018(12 Leads) & AFDB(2 Leads) & SPHDB(1 Lead) \\
\midrule
DCGCNet    & 0.9967 & 0.9951 & 0.9947 \\
w/o AC     & 0.9792 & 0.9715 & 0.9624 \\
w/o ACVQ   & 0.9658 & 0.9543 & 0.9431 \\
\bottomrule
\end{tabular}
\end{table}

\begin{table}[h]
\footnotesize
\setlength{\tabcolsep}{3pt}
\caption{Ablation study on robustness to common ECG noises (ratio=0.333) —BLW, PLI, and EMG—within the intra-dataset.}
\label{tab:ablation3}
\begin{tabular}{lcccccc}
\toprule
 & \multicolumn{3}{c}{Chapman} & \multicolumn{3}{c}{CPSC2018} \\
\cmidrule(lr){2-4} \cmidrule(lr){5-7}
Setting & BLW & PLI & EMG & BLW & PLI & EMG \\
\midrule
DCGCNet    & 0.9941 & 0.9952 & 0.9889 & 0.9863 & 0.9914 & 0.9807 \\
w/o ACVQ   & 0.9624 & 0.9681 & 0.9432 & 0.9415 & 0.9523 & 0.9218 \\
w/o LGCM\_L& 0.9832 & 0.9847 & 0.9603 & 0.9684 & 0.9721 & 0.9456 \\
w/o LGCM\_G& 0.9756 & 0.9802 & 0.9714 & 0.9592 & 0.9667 & 0.9523 \\
w/o LGCM   & 0.9518 & 0.9584 & 0.9276 & 0.9237 & 0.9341 & 0.9024 \\
\bottomrule
\end{tabular}
\end{table}

\subsection{Visualization}
\begin{figure*}
	\centering
	\includegraphics[width=1.0\textwidth]{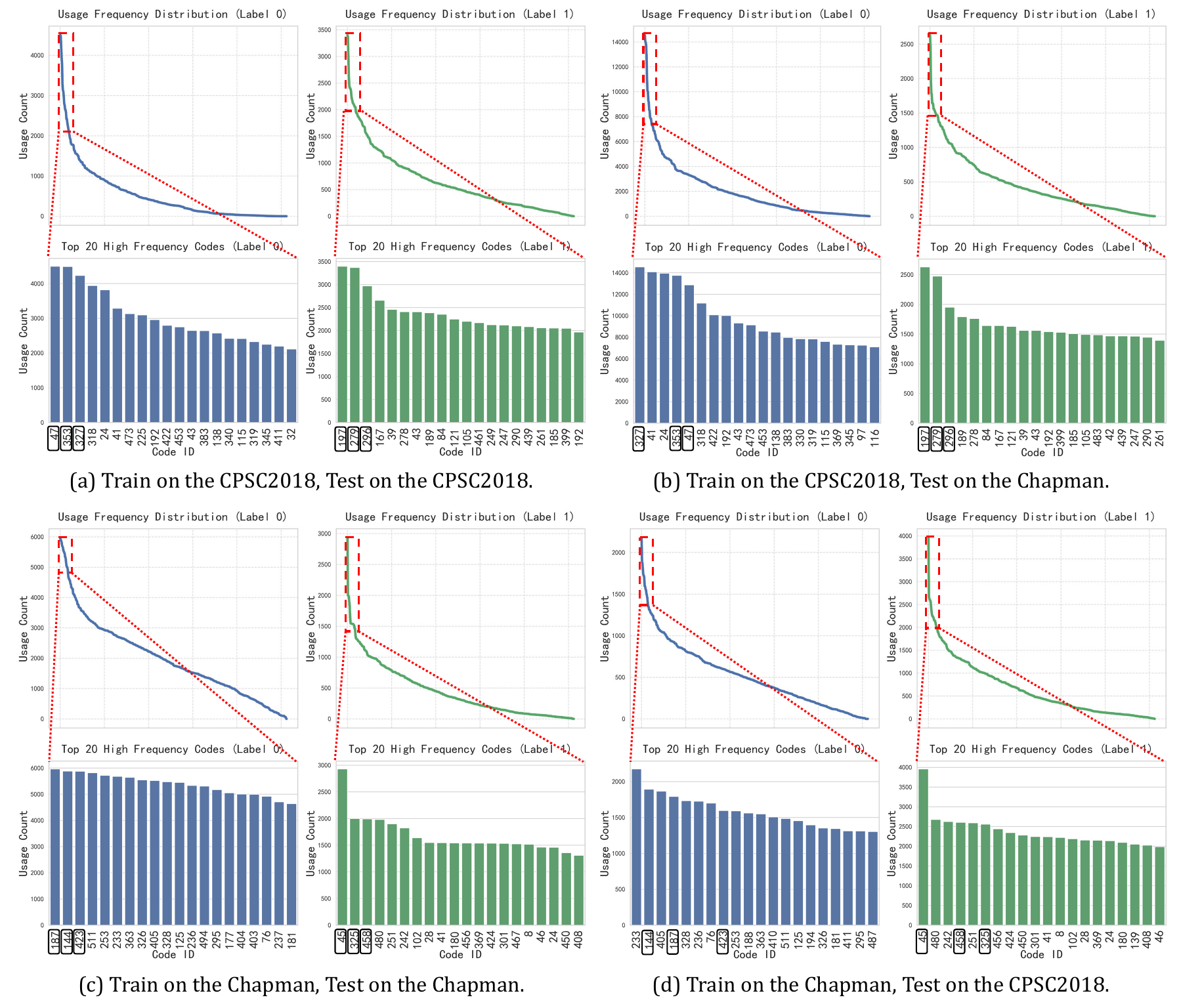}
	\caption{
        Codebook usage patterns across training and testing domains reveal consistent, label-specific prototypes, demonstrating the generalizability of learned discrete representations. 
(a) Model trained on CPSC2018 and tested on CPSC2018; 
(b) Same model tested on Chapman (out-of-domain). 
(c) Model trained on Chapman and tested on Chapman; 
(d) Same model tested on CPSC2018 (out-of-domain). 
        }
	\label{fig:codebook}
\end{figure*}

To investigate the generalization capability of the learned discrete morphological codebook, we visualized the usage frequency of individual codebook vectors across different classes and datasets. As shown in Figure~\ref{fig:codebook}, we compared intra-domain and cross-domain evaluation settings using ECG databases: CPSC2018 and Chapman. When the model was trained on CPSC2018 (Fig.~\ref{fig:codebook}a–b), the most frequently activated codebook entries for class 0 (e.g., indices 47, 353, 327) and class 1 (e.g., 197, 279, 296) remained highly consistent between the in-domain test set (CPSC2018) and the out-of-domain test set (Chapman). Similarly, a model trained on Chapman (Fig.~\ref{fig:codebook}c–d) exhibited stable usage of class-specific codebook vectors—such as indices 187, 144, and 423 for class 0, and 45, 325, 458 for class 1—across both its native test set and the CPSC2018 dataset. This consistency under distribution shift strongly suggests that the codebook does not merely memorize dataset-specific artifacts but instead encodes class-discriminative morphological primitives that are shared across populations and recording protocols. In addition, we presented Grad-CAM interpretability analysis in the Figure\ref{fig:gradcam}.

\section{Conclusion}
In this work, we present DCGCNet, the first codebook-based joint learning framework for atrial fibrillation (AF) detection that simultaneously optimizes classification accuracy and ECG signal reconstruction. By introducing a rhythm codebook to capture local discriminative patterns and a morphology codebook to preserve global waveform fidelity, our model enables synergistic representation learning. Extensive experiments demonstrate state-of-the-art performance across diverse and challenging scenarios, including intra-dataset 12-lead evaluation, cross-dataset arbitrary-lead generalization, and noisy real-world conditions, highlighting its robustness and clinical applicability. Furthermore, we provide multi-level interpretability: statistical analysis of codebook usage reveals consistent activation patterns across domains, offering insight into the model’s generalization mechanism; while Grad-CAM visualizations confirm that diagnostic decisions are grounded in clinically relevant ECG segments. These attributes collectively advance trustworthy AI for AF detection. Future work will explore extending the framework to multi-class arrhythmia diagnosis and real-time deployment on edge devices.

% Conflicts of Interest
\section*{Declaration of Competing Interests}
The authors declare no conflicts of interest.

% Funding
\section*{Funding Sources}
This research received no specific grant.

% Declaration of Generative AI Use
\section*{Declaration of Generative AI Use}
The authors confirm that this manuscript was prepared with the assistance of the Qwen3 large language model for language polishing and formatting. The use of this AI tool did not influence the scientific content, data analysis, or interpretation of results presented in this work. All authors have reviewed and approved the final version of the manuscript.

% Data Statement
\section*{Data Statement}
The following publicly available datasets were used in this study:
Chapman:  \url{https://doi.org/10.13026/wgex-er52};
CPSC2018: \url{http://2018.icbeb.org/Challenge.html};
AFDB: \url{https://www.physionet.org/content/afdb/1.0.0/};
LTAFDB: \url{https://physionet.org/content/ltafdb/1.0.0/};
CPSC2021: \url{https://doi.org/10.13026/ksya-qw89};
SPHDB: \url{https://doi.org/10.17632/dvb5mnhfc4.1};
SHDBAF: \url{https://doi.org/10.13026/n6yq-fq90}.
The source code is publicly available at \url{https://github.com/Ou-Young-1999/DCGCNet}.

\section*{Author Contributions}
Conceptualization: H. Li, X. Ouyang;
Data curation: H. Li, J. Xiao, Y. Lai, S. Lv;
Formal analysis: H. Li, X. Ouyang, J. Wei;
Investigation: H. Li, X. Ouyang, J. Xiao, Y. Lai, S. Lv;
Methodology: H. Li, X. Ouyang, J. Wei;
Project administration: H. Li, X. Ouyang, J. Xiao;
Resources: H. Li, J. Xiao;
Software: X. Ouyang;
Supervision: H. Li;
Validation: H. Li, X. Ouyang, J. Wei, G. Li, Y. Lei, G. Ma;
Visualization: X. Ouyang, Y. Lei;
Writing – original draft: X. Ouyang;
Writing – review \& editing: H. Li, X. Ouyang, J. Wei, G. Li, Y. Lei, G. Ma.

\printcredits

\bibliographystyle{cas-model2-names}
% \bibliographystyle{elsarticle-num}

\bibliography{cas-refs}

\appendix
\section{My Appendix}
\renewcommand{\thefigure}{\thesection.\arabic{figure}}   % 图编号：A.1, A.2...
\renewcommand{\thetable}{\thesection.\arabic{table}}     % 表编号：A.1, A.2...
\setcounter{figure}{0}  % 重置图编号
\setcounter{table}{0}   % 重置表编号
\begin{figure*}[h]
	\centering
	\includegraphics[width=1.0\textwidth]{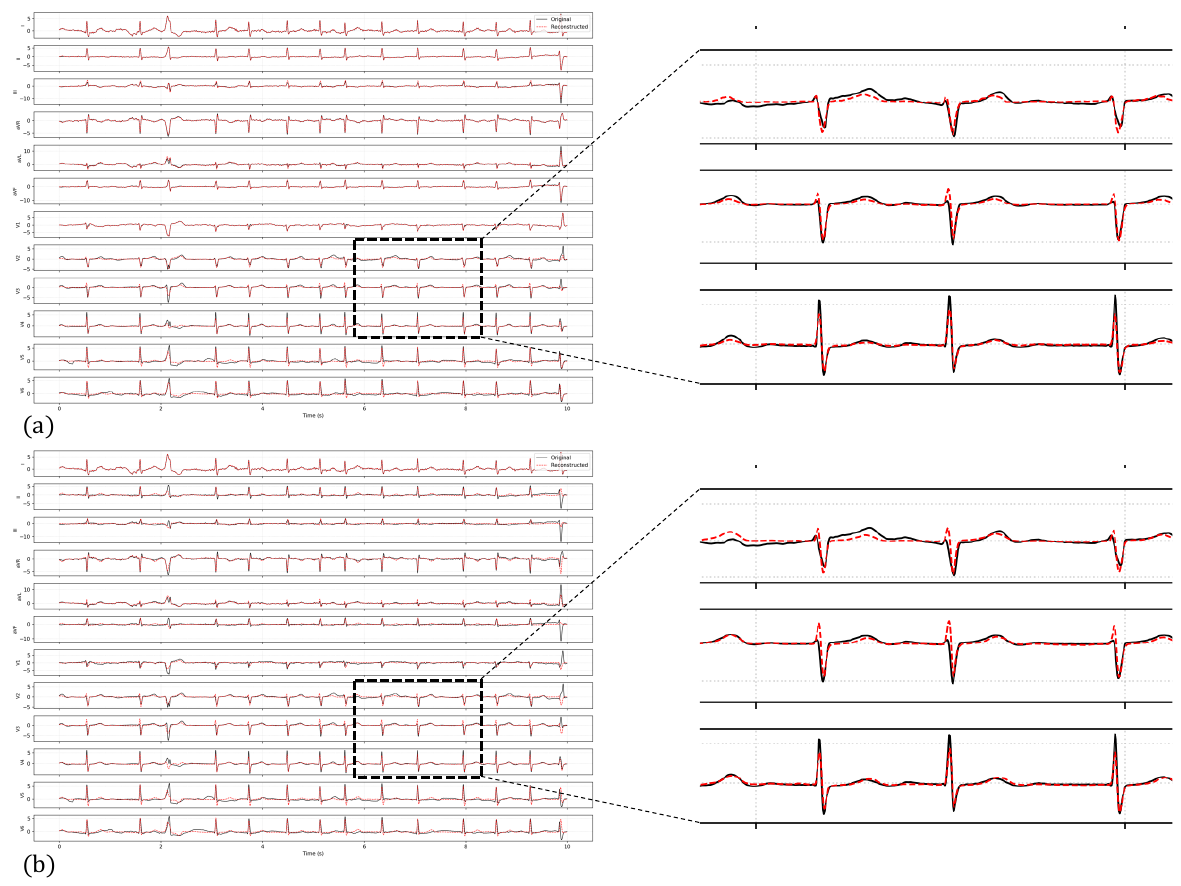}
	\caption{
        12-lead ECG reconstruction results generated by our proposed DCGCNet. 
        (a) Reconstruction using the combination of Leads I, II, and V1 as input. 
        (b) Reconstruction using only Lead I as input. 
        In both subplots, the ground-truth 12-lead ECG signals are plotted as solid black lines, while the DCGCNet-predicted reconstructions are shown as dashed red lines.
        }
	\label{fig:reconstruction}
\end{figure*}

\begin{figure*}[h]
	\centering
	\includegraphics[width=1.0\textwidth]{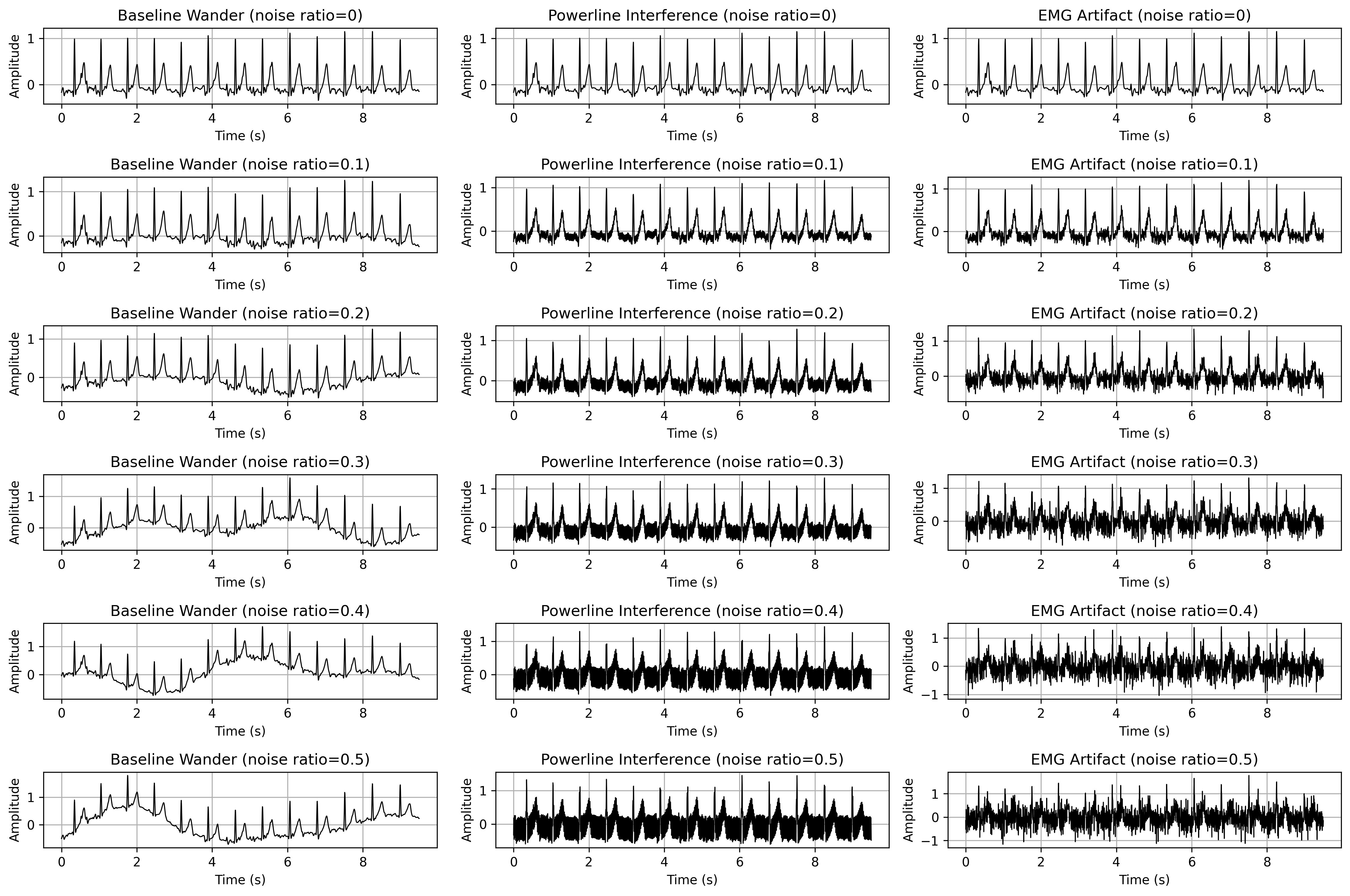}
	\caption{
        Effect of clinically relevant additive noise on the original 12-lead ECG signal. Three common artifacts—baseline wander (caused by respiration or motion, simulated as low-frequency sinusoids <0.3 Hz), powerline interference (from mains electricity, modeled as a 50 Hz sine wave with random phase), and electromyographic (EMG) artifact (due to muscle activity, generated as bandpass-filtered white noise in 20–250 Hz)—are injected at intensity ratios of 0.1–0.5 relative to the ECG’s peak-to-peak amplitude, ensuring physiologically realistic corruption.
        }
	\label{fig:noise}
\end{figure*}

\begin{table*}[h]
\centering
\caption{Comparison of AF detection performance with state-of-the-art methods across multiple datasets and cross-dataset settings. All results are directly reported from the original publications, ensuring a faithful reproduction of published claims. Metrics are presented as percentages (\%). Abbreviations: CV = cross-validation; TL = transfer learning; CL = contrastive learning; DTW = dynamic time warping; AE = autoencoder; HN = hierarchical normalization. The proposed method (``Our'') demonstrates consistently competitive or superior performance, particularly in challenging cross-dataset generalization scenarios.}
\label{tab:sota_comparison}
\begin{tabular}{lllccccccll}
\toprule
\textbf{Train} & \textbf{Test} & \textbf{Input} & \textbf{AUC} & \textbf{ACC} & \textbf{F1} & \textbf{Precision} & \textbf{Recall} & \textbf{Method} & \textbf{Other} \\
\midrule
AFDB & AFDB & 30s & 99.50 & 97.10 & -- & -- & 97.00 & Wavelet + SVM\cite{svm} & 2-fold CV \\
AFDB & AFDB & 30 beats & -- & 97.60 & 97.10 & -- & 98.00 & Peak + RF\cite{rf} & 4-fold CV \\
AFDB & AFDB & 1.2s & -- & 99.23 & -- & -- & 99.41 & Wavelet + CNN\cite{cnns} & 4:1 split \\
AFDB & AFDB & 30s & -- & 97.80 & -- & -- & 98.98 & CNN + LSTM\cite{cnnlstm} & 5-fold CV \\
AFDB & AFDB & 30s & -- & 93.00 & 92.64 & -- & 89.83 & Siamese + Few-shot TL\cite{fewshottl} & 5-fold CV \\
AFDB & AFDB & 15s & -- & 99.11 & 97.00 & 99.46 & 95.40 & Spatio-temporal Siamese\cite{siamese} & 8:1:1 split \\
AFDB & AFDB & 5s & -- & 99.60 & -- & -- & 99.43 & Transformer\cite{selfattention} & 95\%/5\%/5\% \\
Wearable I & AFDB & 10s & -- & 95.28 & -- & -- & 96.46 & LSTM + CNN\cite{lstmcnn} & 10-fold CV \\
CPSC2021 & AFDB & 30s & 99.53 & 98.63 & 98.28 & 99.23 & 97.35 & ResCNN + BiLSTM\cite{rescnnbilstm} & 5-fold CV \\
CPSC2021 & AFDB & 30s & -- & 98.18 & 98.10 & 98.33 & 97.90 & Semi-supervised CL\cite{cl} & Cross-dataset \\
Chapman & AFDB & 10s & \textbf{99.51} & \textbf{97.55} & \textbf{97.40} & \textbf{97.30} & \textbf{97.49} & \textbf{Ours} & 5-fold CV \\
\midrule

CPSC2021 & CPSC2021 & 12 beats & -- & 95.19 & -- & -- & 96.78 & DTW + AE + SVM\cite{dtw} & 5-fold CV \\
CPSC2021 & CPSC2021 & 30s & 99.70 & 98.63 & 98.18 & 97.34 & 99.08 & ResCNN + BiLSTM\cite{rescnnbilstm} & 5-fold CV \\
CPSC2021 & CPSC2021 & 5s & -- & 98.40 & -- & -- & 98.42 & Transformer\cite{selfattention} & 95\%/5\%/5\% \\
UVAF & CPSC2021 & 60 beats & 99.00 & -- & 95.00 & -- & 95.00 & ResNet + GRU\cite{resnetgru} & 5-fold CV \\
Chapman & CPSC2021 & 10s & \textbf{98.96} & \textbf{96.28} & \textbf{95.96} & \textbf{95.37} & \textbf{96.68} & \textbf{Ours} & 5-fold CV \\
\midrule

AFDB & LTAF & 30s & -- & 97.09 & 97.07 & -- & 97.37 & Siamese + Few-shot TL\cite{fewshottl} & 5-fold CV \\
CPSC2021 & LTAF & 30s & 98.29 & 97.07 & 97.23 & 97.66 & 96.81 & ResCNN + BiLSTM\cite{rescnnbilstm} & 5-fold CV \\
CPSC2021 & LTAF & 30s & -- & 96.61 & 96.60 & 96.56 & 96.64 & Semi-supervised CL\cite{cl} & Cross-dataset \\
Chapman & LTAF & 10s & \textbf{99.04} & \textbf{96.44} & \textbf{96.42} & \textbf{96.40} & \textbf{96.45} & \textbf{Ours} & 5-fold CV \\
\midrule

UVAF & SHDBAF & 60 beats & 99.00 & -- & 92.00 & -- & 93.00 & ResNet + GRU\cite{resnetgru} & 5-fold CV \\
Chapman & SHDBAF & 10s & \textbf{98.78} & \textbf{96.30} & \textbf{94.17} & \textbf{93.75} & \textbf{94.63} & \textbf{Ours} & 5-fold CV \\
\midrule

Chapman & Chapman & 10s & 99.80 & 98.40 & 97.70 & -- & -- & Supervised CL + HN\cite{norm} & 8:1:1 split \\
Chapman & Chapman & 10s & \textbf{99.98} & \textbf{99.66} & \textbf{99.43} & \textbf{99.24} & \textbf{99.62} & \textbf{Ours} & 5-fold CV \\
CPSC2018 & Chapman & 10s & \textbf{98.74} & \textbf{94.45} & \textbf{91.69} & \textbf{88.44} & \textbf{96.58} & \textbf{Ours} & 5-fold CV \\
\midrule

CPSC2018 & CPSC2018 & 10s & 99.50 & 94.40 & 94.40 & -- & -- & Supervised CL + HN\cite{norm} & 8:1:1 split \\
CPSC2018 & CPSC2018 & 10s & \textbf{99.91} & \textbf{99.37} & \textbf{99.36} & \textbf{99.32} & \textbf{99.41} & \textbf{Ours} & 5-fold CV \\
Chapman & CPSC2018 & 10s & \textbf{99.67} & \textbf{96.29} & \textbf{96.27} & \textbf{96.14} & \textbf{96.67} & \textbf{Ours} & 5-fold CV \\

\bottomrule
\end{tabular}
\end{table*}

\begin{table*}[h]
\centering
\caption{Sensitivity analysis of model performance to auxiliary loss weights, with $\lambda_{\text{cls}}=1$ and $\lambda_{\text{rec}}=1$ fixed. Reported metrics are AUC and accuracy (ACC) on Chapman and CPSC2018 datasets.}
\label{tab:sensitivity}
\begin{tabular}{lcccccccc}
\toprule
Setting & $\lambda_{\text{vq}}$ & $\lambda_{\text{cong}}$ & $\lambda_{\text{conl}}$ &
Chapman AUC & Chapman ACC & CPSC2018 AUC & CPSC2018 ACC \\
\midrule
Default          & 0.2  & 0.5  & 0.1  & 99.98 & 99.66 & 99.91 & 99.37 \\
Weaker global    & 0.2  & 0.1  & 0.1  & 99.74 & 98.83 & 99.48 & 98.26 \\
Stronger global  & 0.2  & 1.0  & 0.1  & 99.80 & 99.03 & 99.62 & 98.63 \\
Excessive global & 0.2  & 3.0  & 0.1  & 99.22 & 97.59 & 98.51 & 96.54 \\
Weaker local     & 0.2  & 0.5  & 0.02 & 99.66 & 98.54 & 99.35 & 98.07 \\
Stronger local   & 0.2  & 0.5  & 0.20 & 99.70 & 98.75 & 99.54 & 98.42 \\
Excessive local  & 0.2  & 0.5  & 0.50 & 99.11 & 97.20 & 98.34 & 96.29 \\
Weaker VQ        & 0.05 & 0.5  & 0.1  & 99.51 & 98.24 & 99.08 & 97.53 \\
Stronger VQ      & 0.50 & 0.5  & 0.1  & 99.34 & 97.83 & 98.76 & 97.05 \\
Excessive VQ     & 1.00 & 0.5  & 0.1  & 98.59 & 96.20 & 97.67 & 95.28 \\
\bottomrule
\end{tabular}
\end{table*}

\begin{figure*}[h]
	\centering
	\includegraphics[width=0.9\textwidth]{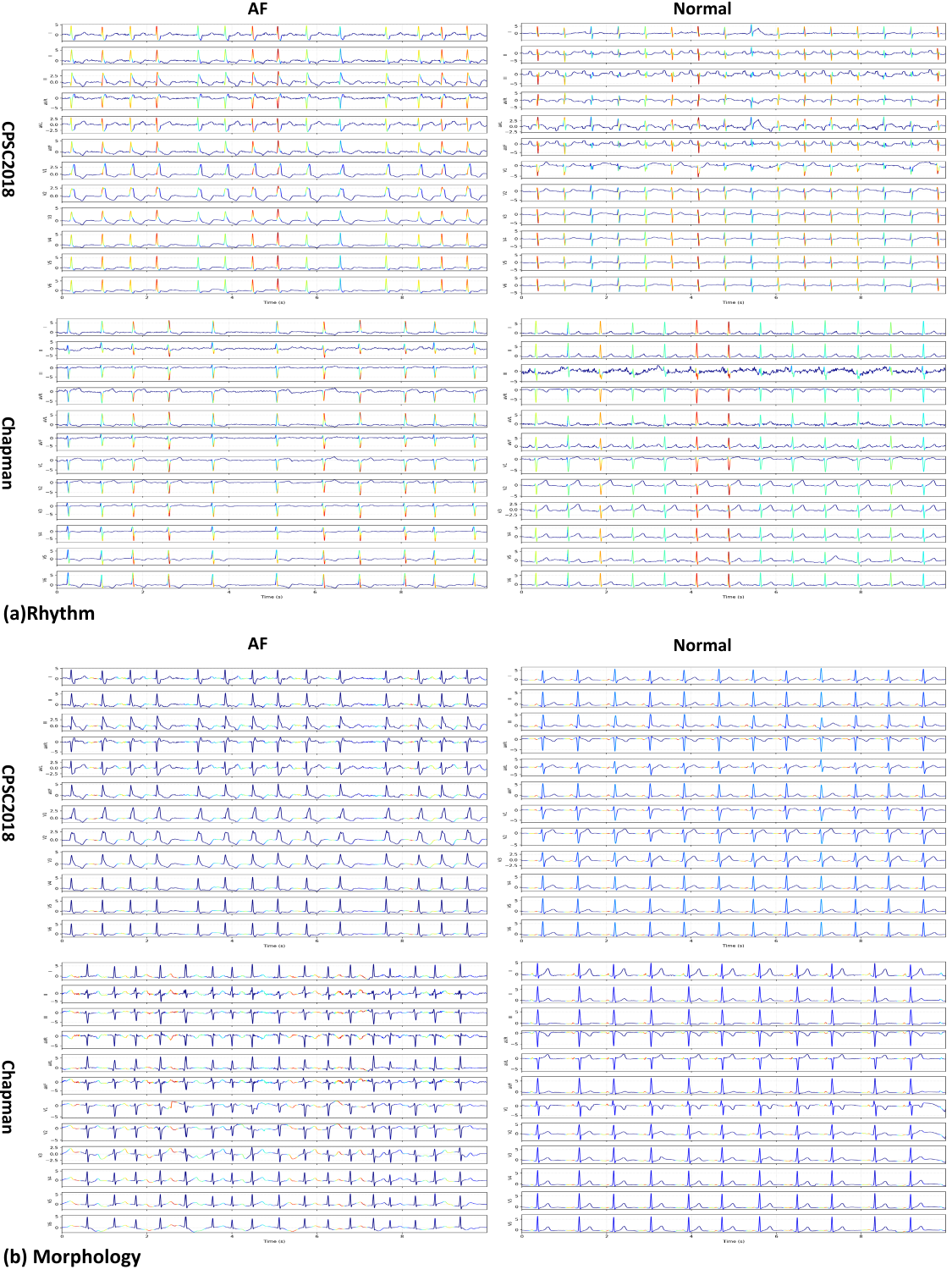}
	\caption{
        Grad-CAM visualizations validate the functional specialization of our dual-encoder architecture. Color intensity reflects gradient-based importance: warm colors (red/yellow) indicate regions most influential to each encoder’s output, while cool colors (blue) denote low attention. (a) The rhythm encoder—using small kernels—consistently attends to the R-wave, its interval regularity directly underpins rhythm assessment and atrial fibrillation detection. (b) The morphology encoder—employing larger kernels—focuses on the P-wave region, which carries structural information about atrial depolarization: organized P-waves signify sinus rhythm, whereas their absence or disorganization is characteristic of atrial fibrillation. This clear dissociation confirms our design rationale: distinct receptive fields effectively decouple temporal dynamics (rhythm) from waveform structure (morphology).  
        }
	\label{fig:gradcam}
\end{figure*}


% Author biography without author photo.
% Author biography. Author biography. Author biography.
% Author biography. Author biography. Author biography.
% Author biography. Author biography. Author biography.
% Author biography. Author biography. Author biography.
% Author biography. Author biography. Author biography.
% Author biography. Author biography. Author biography.
% Author biography. Author biography. Author biography.
% Author biography. Author biography. Author biography.
% Author biography. Author biography. Author biography.

% \bio{figs/cas-pic1}
% Author biography with author photo.
% Author biography. Author biography. Author biography.
% Author biography. Author biography. Author biography.
% Author biography. Author biography. Author biography.
% Author biography. Author biography. Author biography.
% Author biography. Author biography. Author biography.
% Author biography. Author biography. Author biography.
% Author biography. Author biography. Author biography.
% Author biography. Author biography. Author biography.
% Author biography. Author biography. Author biography.

% \bio{figs/cas-pic1}
% Author biography with author photo.
% Author biography. Author biography. Author biography.
% Author biography. Author biography. Author biography.
% Author biography. Author biography. Author biography.
% Author biography. Author biography. Author biography.

\end{document}